\documentclass[10pt,twocolumn,letterpaper]{article}

\usepackage{iccv}              

\usepackage{graphicx}  
\usepackage{subcaption} 

\usepackage[dvipsnames]{xcolor}
\usepackage{colortbl}

\definecolor{orange}{rgb}{1, 0.5, 0.}
\definecolor{darkgreen}{rgb}{0, 0.3, 0}
\definecolor{green}{rgb}{0, 0.6, 0}

\definecolor{placeholder}{rgb}{0.6,0.8,0.95}

\definecolor{amber}{rgb}{1.0, 0.75, 0.0}

\definecolor{visible-blue}{rgb}{0.286, 0.525, 0.910}

\definecolor{tabfirst}{rgb}{1, 0.7, 0.7} 
\definecolor{tabsecond}{rgb}{1, 0.85, 0.7} 
\definecolor{tabthird}{rgb}{1, 1, 0.7} 
\usepackage{overpic}
\usepackage{wrapfig}
\usepackage{amsmath}
\usepackage{empheq}
\usepackage{comment}
\usepackage{makecell}
\usepackage{multirow}
\usepackage{multicol}
\usepackage{float}
\usepackage{marvosym}
\usepackage{amsmath}
\usepackage{booktabs}
\usepackage{xcolor}
\definecolor{iccvblue}{rgb}{0.21,0.49,0.74}
\usepackage[pagebackref,breaklinks,colorlinks,allcolors=iccvblue]{hyperref}

\def\paperID{664} 
\def\confName{ICCV}
\def\confYear{2025}

\title{NeRF Is a Valuable Assistant for 3D Gaussian Splatting}

\author{Shuangkang Fang\textsuperscript{1}, 
~I-Chao Shen\textsuperscript{2}, 
~Takeo Igarashi\textsuperscript{2}, 
~Yufeng Wang\textsuperscript{1}, 
~ZeSheng Wang\textsuperscript{1},\\ 
~Yi Yang\textsuperscript{3}, 
~Wenrui Ding\textsuperscript{1}, 
~Shuchang Zhou\textsuperscript{3}\\
\textsuperscript{1}Beihang University~~~ \textsuperscript{2}The University of Tokyo~~~
\textsuperscript{3}StepFun\\
}

\newcommand{\methodname}{NeRF-GS}

\begin{document}
\maketitle
\begin{abstract}
    We introduce \methodname{}, a novel framework that jointly optimizes Neural Radiance Fields (NeRF) and 3D Gaussian Splatting (3DGS). This framework leverages the inherent continuous spatial representation of NeRF to mitigate several limitations of 3DGS, including sensitivity to Gaussian initialization, limited spatial awareness, and weak inter-Gaussian correlations, thereby enhancing its performance. 
    In \methodname{}, we revisit the design of 3DGS and progressively align its spatial features with NeRF, enabling both representations to be optimized within the same scene through shared 3D spatial information. We further address the formal distinctions between the two approaches by optimizing residual vectors for both implicit features and Gaussian positions to enhance the personalized capabilities of 3DGS. Experimental results on benchmark datasets show that \methodname{} surpasses existing methods and achieves state-of-the-art performance. This outcome confirms that NeRF and 3DGS are complementary rather than competing, offering new insights into hybrid approaches that combine 3DGS and NeRF for efficient 3D scene representation.
\end{abstract}    
\section{Introduction}
\label{sec:intro}

\begin{figure}[t]
  \centering
   \includegraphics[width=0.9\linewidth]{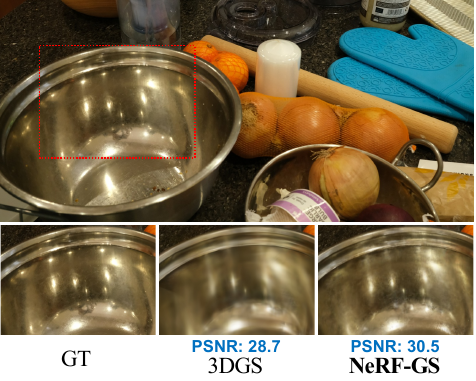}
   \caption{\methodname{} establishes a bridge of communication between NeRF and 3DGS, leveraging information sharing, modeling of distinct characteristics, and joint optimization to enable 3DGS to achieve higher fidelity representation. In this case, \methodname{} outperforms the vanilla 3DGS by 1.8dB in PSNR.}
   \label{fig:teaser}
\end{figure}

 Neural Radiance Fields (NeRF)~\cite{mildenhall2020nerf} and 3D Gaussian Splatting (3DGS)~\cite{kerbl2023-3dgs} have emerged as two prominent methodologies in 3D scene reconstruction, photorealistic rendering, and virtual reality applications~\cite{mildenhall2020nerf, mueller2022instant, fang2023pvdal, fridovich2022plenoxels, chen2022tensorf, fang2024-ce3d, wang2021neus, martinbrualla2020nerfw, fang2023dn2n, samavati20233D-rec-survey, picard2023survey, fang2024sff}.
 NeRF represents 3D scenes through a continuous volumetric field, capturing intricate details by an MLP-based encoding of color and density at any position in space. However, it requires numerous forward passes through the MLP, limiting its applicability in real-time scenarios.
In contrast, 3DGS~\cite{kerbl2023-3dgs} represents a scene through a set of discrete Gaussians to approximate points in space, which capitalizes on point-based rendering for computational efficiency, providing real-time performance. Nonetheless, the reliance on Gaussian initialization and limited spatial perception can lead to instability in 3DGS training~\cite{fan2024instantsplat, colmapfree3dgs, niemeyer2024radsplat, foroutan2024-NerfInit}. Moreover, the weak correlation between discrete Gaussians results in a lack of smooth spatial transitions~\cite{mihajlovic2025splatfields, chen2025hac, chen2024survey-3dgs}, which negatively affects the visual quality of the rendered outputs.

To address these deficiencies, existing studies have sought to improve both NeRF and 3DGS. For example, some researches focus on accelerating rendering for NeRF~\cite{mueller2022instant,chen2022tensorf,yu2021plenoctrees,fridovich2022plenoxels,reiser2021kilonerf,garbin2021fastnerf}, while others improve 3DGS in terms of 
visual quality~\cite{mihajlovic2025splatfields,lu2024scaffold,yu2024mip-gs,chen2025mvsplat}. However, most of these studies treat NeRF and 3DGS as independent scene representation paradigms, concentrating on their separate enhancements. 
Several researchers have attempted to leverage the properties of NeRF to enhance 3DGS, such as initializing 3DGS with NeRF~\cite{foroutan2024-NerfInit,niemeyer2024radsplat}, embedding NeRF attributes within 3DGS~\cite{malarz2023-vdgs,mihajlovic2025splatfields,niemeyer2024radsplat}, or creating networks to implicitly estimate 3DGS parameters~\cite{lu2024scaffold,chen2025hac}.
However, these approaches primarily focus on individually enhancing variants of 3DGS, without systematically exploring the potential components in the full NeRF pipeline that could benefit 3DGS. They also overlook the modeling of structural differences between the two, resulting in limited performance gains. Thus, the integration of NeRF and 3DGS methods and the combination of their respective strengths remain underexplored.

To this end, we propose \methodname{}, a novel framework that integrates the NeRF network into the training of the 3DGS model, leveraging specific NeRF properties to address 3DGS inherent limitations. In revisiting the design space of the 3DGS model, we identify and implement three critical components in the hybrid \methodname{} framework.

(1) Sharing Mechanism (Sec~\ref{subsec:share}): we first introduce a Hash-based network for encoding features in continuous space optimized by NeRF volume rendering, and design strategies to identify potential Gaussian positions. Subsequently, both NeRF and 3DGS share these features to decode additional attributes for their respective spatial points.

(2) Residual Vectors (Sec~\ref{subsec:div}): due to inherent differences between the NeRF and 3DGS forms, directly using NeRF-optimized features and NeRF-initialized Gaussian positions does not adequately adapt to the 3DGS branch. To address this, we propose explicitly modeling their discrepancies by optimizing residual vectors for both features and positions to personalize and enhance 3DGS performance.

(3) Joint Optimization (Sec~\ref{subsec:joint}): we align the attributes and rendering results of spatial points along rays passing through the important Gaussian in the NeRF branch with those in the 3DGS, which reduces feature confusion and ensures mutually beneficial constraints on shared features. Additionally, we leverage NeRF's continuous spatial query capability to assist in adaptive Gaussian growth, achieving efficient joint optimization across different branches.

The hybrid design of \methodname{} is not a mere combination of NeRF and 3DGS, but rather a systematic and comprehensive integration that considers the interrelations and differences between the two, maximizing the auxiliary role of NeRF in enhancing 3DGS. 
It is structurally flexible, allowing the 3DGS branch to be independently separated after joint optimization, thus preserving its real-time rendering capability. Experiments conducted on benchmark datasets demonstrate that \methodname{} significantly outperforms the original 3DGS method in both quantitative and qualitative evaluation. 
Additionally, the mutual regularization between the dual branches in \methodname{} notably improves the rendering quality of the 3DGS branch under sparse-view conditions. These findings indicate that these seemingly disparate scene representation methods are, in fact, complementary rather than competitive, providing new insights for exploring further hybrid 3D scene representation techniques.
\section{Related work}
\label{sec:related_work}

\noindent\textbf{Implicit Volume Rendering.} Implicit methods provide a continuous spatial modeling capability to represent 3D scenes, eliminating the need for discretization~\cite{mildenhall2020nerf,mueller2022instant,barron2021mip,zhang2022ray,atzmon2019controlling,mescheder2019occupancy,michalkiewicz2019implicit,niemeyer2019occupancy,oechsle2019texture,park2019deepsdf,peng2020convolutional,lindell2021autoint,sun2022direct}. Many methods have been developed on this basis to improve the visual quality and rendering speed. Plenoctrees~\cite{yu2021plenoctrees} and Plenoxels~\cite{fridovich2022plenoxels}  render faster than vanilla NeRF by pre-tabulating a Tensor. 
DeRF \cite{rebain2021derf} and KiloNeRF \cite{reiser2021kilonerf} accelerate speed by partitioning the target scene into smaller MLPs.
Instant-NGP \cite{mueller2022instant} introduces a learnable, multi-resolution hash encoding to fit scenes efficiently. 
Mip-NeRF \cite{barron2021mip} enhances NeRF with cone tracing multi-scale properties and automatic anti-aliasing. 
Several methods have demonstrated that the features extracted by NeRF contain significant scene information. For example, Unisurf~\cite{oechsle2021unisurf} achieves a detailed mesh by sharing features between NeRF and SDF. DecomNeRF~\cite{decomposing-nerf} enables semantic-level scene decomposition through feature embedding. PVD~\cite{fang2023pvd,fang2023pvdal} facilitates the conversion between different forms of NeRF by distilling features.

\vspace{1mm}
\noindent\textbf{Point-based Representations.}
Recent advances in point-based 3D rendering have shown substantial improvements in rendering efficiency~\cite{kerbl2023-3dgs,malarz2023-vdgs,lu2024scaffold,huang20242dgs,jiang2024gaussianshader,gao2023relightable,mihajlovic2025splatfields,jung2024relaxing,yang2024gaussianobject}. 
RAIN-GS~\cite{jung2024relaxing}, Agg~\cite{xu2024agg}, and NPGs~\cite{das2024NPGs} have proposed novel initialization strategies to address the limitations of the initialization from SfM in the original 3DGS. 
MS3DGS~\cite{yan2024multi}, Analytic-Splatting~\cite{liang2024analytic}, and SA-GS~\cite{song2024sa} enhance 3DGS performance by introducing strategies to reduce aliasing. Additionally, to mitigate the issues of storage demands in 3DGS, some approaches have achieved lightweight Gaussian representations through parameter compression~\cite{niedermayr2024-c3dgs, zeghidour2021soundstream, navaneet2023compact3d,bag20243dgs-comp-survey,liu2024comp-gs} and pruning~\cite{fan2023lightgaussian,zhang2024lp-prun1,ali2024trimming-prun2}.

Several studies have explored the complementarity and transfer of characteristics between different 3D representations. 
Notable examples~\cite{foroutan2024-NerfInit, niemeyer2024radsplat} utilize points extracted from NeRF for 3DGS initialization. VDGS~\cite{malarz2023-vdgs} incorporates NeRF concepts by employing implicit MLPs to make 3DGS opacity view-dependent. SplatFields~\cite{mihajlovic2025splatfields} samples implicit features from triplanes, establishing an auto-correlated feature space for estimating Gaussian sphere parameters.
Scaffold-GS~\cite{lu2024scaffold} derives the possible positions and attributes of Gaussian from a set of candidate anchors. Hash-GS~\cite{chen2025hac} and Compact-3DGS~\cite{lee2024compact-3dgs} leverages NeRF attributes for 3DGS parameter compression.
However, these methods mainly adopt certain NeRF-inspired features to independently optimize 3DGS variants, which differ significantly from our methodology. Moreover, these methods typically implement direct transfer of NeRF characteristics to 3DGS without considering their inherent differences, thereby failing to fully exploit the models' potential.

\section{Preliminaries}
\label{sec:preliminaries}

\begin{figure*}[t]
  \centering
\includegraphics[width=1\linewidth]{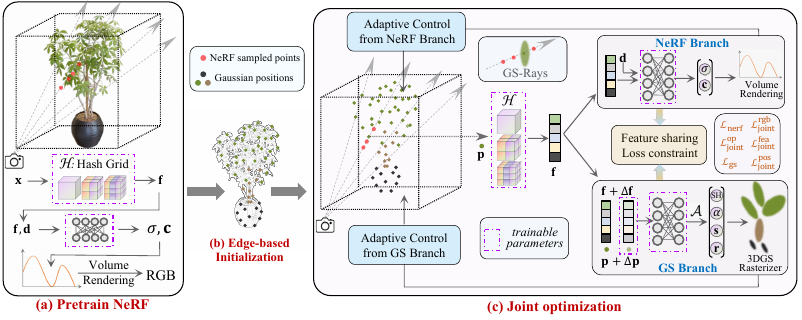}
   \caption{\textbf{Overview of \methodname}. (a) We first pretrain a Hash-based NeRF network to acquire continuous spatial encoding capabilities and implicit scene representation.
(b) Utilizing the preliminary scene carved by NeRF, we resample rays corresponding to image edges to obtain potential Gaussian positions, facilitating Gaussian initialization.
(c) During joint optimization, the GS branch queries corresponding features $\mathbf{f}$ from the Hash grid $\mathcal{H}$ for each Gaussian sphere. These features, combined with positions $\mathbf{p}$ and their respective residual terms ($\Delta \mathbf{f}, \Delta \mathbf{p}$), decode additional Gaussian attributes $\mathcal{A}$, including color, opacity, scale, and rotation vectors. For the NeRF branch, rendering is exclusively performed on rays (GS-Rays) passing through important Gaussian spheres within the view frustum. The two branches are aligned by opacity and RGB values ($\mathcal{L}_{\text{joint}}^\text{op}, \mathcal{L}_{\text{joint}}^\text{rgb}$), further supervised by reconstruction($\mathcal{L}_{\text{gs}},\mathcal{L}_{\text{nerf}} $), and residual regularization ($\mathcal{L}_{\text{reg}}^\text{fea}, \mathcal{L}_{\text{reg}}^\text{pos}$). Simultaneously, we leverage ray attributes from the NeRF branch along with gradient information from the GS branch to achieve adaptive control over Gaussian spheres. The purple dashed box marks the parameters to be trained.}
   \label{fig:pipeline}
\end{figure*}

\noindent\textbf{Neural Radiance Fields.}
NeRF represents scenes using an implicit function that maps spatial points \(\mathbf{x} = (x, y, z)\) and view directions \(\mathbf{d} = (\theta, \phi)\) to density \(\sigma\) and color \(\mathbf{c}\). For a ray originating at \(\mathbf{o}\) with direction \(\mathbf{d}\), the RGB value \(C_{\text{nerf}}\) of the corresponding pixel is computed by numerically integrating the colors \(\mathbf{c}_i\) and densities \(\sigma_i\) of the spatial points \(\mathbf{x}_i = \mathbf{o} + t_i\mathbf{d}\) sampled along the ray:
\begin{equation}
    C_{\text{nerf}} = \sum_{i=1}^{N} T_i (1 - \exp(-\sigma_i \delta_i)) \mathbf{c}_i,
    \label{eq:neural_rendering_equation}
\end{equation}
where $T_i = \exp(- \sum_{j=1}^{i-1} \sigma_j \delta_j)$ is the accumulated transmittance up to the \(i\)-th sample, and \(\delta\) is the distance between adjacent samples.

\vspace{1mm}
\noindent\textbf{Gaussian Splatting.} In 3DGS, the scene is represented by a set of anisotropic 3D Gaussian functions, inheriting the EWA volume splatting method \cite{zwicker2001ewa} and allowing efficient rendering through a tile-based rasterization approach. Typically, 3DGS are initialized from a set of points generated by SfM and can be described by the central position $\mathbf{p}$ and a covariance matrix $\Sigma$ that is parameterized using a rotation matrix $R$ and a scaling matrix $S$ as follows:
\begin{equation}
\Sigma = RSS^TR^T.
\label{sigma}
\end{equation}
3DGS uses a quaternion $\mathbf{r}$ to represent rotation and a vector $\mathbf{s}$ for scaling. Each Gaussian is also associated with an opacity $\alpha \in [0,1]$ and a set of spherical harmonics (SH) coefficients to define the view-dependent color $\mathbf{c}$. By projecting the 3D Gaussian into 2D, the color and opacity coverage of each projected Gaussian is evaluated and the pixel color $C_{gs}$ can be calculated by sequentially blending all 2D Gaussians that contribute to the pixel, as follows:
\begin{equation}
    C_{gs} = \sum_{i\in N} \mathbf{c}_i \alpha_i \prod_{j=1}^{i-1}(1-\alpha_j).
    \label{eq:gs_render}
\end{equation}

\section{\methodname}
\label{sec:method}
Our objective is to integrate the full NeRF pipeline into 3DGS model training and utilize specific properties of NeRF to address the limitations of 3DGS. \cref{fig:pipeline} illustrates an overview of our method. 
We start by independently training the NeRF branch to model a spatially continuous Hash feature and initialize the Gaussian spheres in the 3DGS branch, enabling spatial awareness and feature sharing between the two branches (Sec~\ref{subsec:share}). We design a neural GS branch derived from these shared features. To fill the gap between NeRF and 3DGS representations, we further optimize each Gaussian with a residual feature vector and a position offset vector, using the refined vectors to infer additional Gaussian attributes (Sec~\ref{subsec:div}). 
During joint optimization, we introduce `GS-Rays', defined as rays connecting important Gaussian centers within the view frustum to the camera origin, serving as query rays for the NeRF branch. Along these GS-Rays, we achieve mutual constraint between NeRF and GS branches by minimizing differences in their spatial attributes and rendering results. Furthermore, we leverage NeRF to facilitate Gaussian adaptive growth in regions challenging for 3DGS perception (Sec~\ref{subsec:joint}).

\subsection{Sharing Mechanism in Dual-branch}
\label{subsec:share}

\noindent\textbf{NeRF for Prior Sharing.}  
NeRF represents the scene as a continuous volumetric field, allowing arbitrary queries of spatial points to obtain density $\sigma$ and color $\mathbf{c}$. This guarantees that every Gaussian has a corresponding NeRF feature, and volume rendering creates strong spatial correlations that address 3DGS’s limitations in discrete point representation and weak spatial relationships. To achieve efficient feature sharing, we construct a hash feature extraction network $\mathcal{H}$ that extracts multi-scale features $\mathbf{f}$ from a spatial point \(\mathbf{x}\). 
As in INGP~\cite{mueller2022instant}, the density \(\sigma\) is derived from the spatial feature \(\mathbf{f}\), with the color $\mathbf{c}$ derived by combining \(\mathbf{f}\) and the direction vector \(\mathbf{d}\), as shown below.
\begin{equation}
\mathbf{f} = \mathcal{H}(\mathbf{x}), ~~\sigma = F_{\sigma}(\mathbf{f}), ~~\mathbf{c}=F_{c}(\mathbf{f}, \mathbf{d}).
\end{equation}
Once \(\sigma\) and \(\mathbf{c}\) are obtained, the images can be rendered by Eq.~\ref{eq:neural_rendering_equation}. After the NeRF branch has been pre-trained, the features \(\mathbf{f}\) can be shared with the 3DGS branch to capture similar information at corresponding spatial points.

\noindent\textbf{Edge-based Initalization}.
Similar to RadSplat~\cite{niemeyer2024radsplat}, we estimate initial Gaussian positions by computing the median ray depth \( z \). However, unlike RadSplat, which uniformly samples one million points from all rays for Gaussian initialization, we assign higher sampling weights to rays corresponding to high-frequency image textures, as these textures define scene contours. Specifically, we apply edge detection to extract image textures, designate their corresponding rays as edge rays, and then estimate the potential Gaussian position set \( \mathcal{G}_{init} \) using the following approach:
\begin{equation}
\mathcal{G}_{init} = \{\textbf{p}_i ~|~~\textbf{p}_i \in \{\mathcal{P}_{edge}, \mathcal{P}_{random}\} \},
\label{eq:init}
\end{equation}
where $\mathcal{P}_{edge}$ and $\mathcal{P}_{random}$ are points in edge rays and random rays, respectively, and Gaussian position $\textbf{p}_i=\textbf{o} + \textbf{d}_i * z_i$.  
Our design tightly integrates the NeRF and GS branches, ensuring that the initialized points continue to share spatial information and undergo further joint optimization, which is completely different from RadSplat and NeRF-init~\cite{foroutan2024-NerfInit} that treat initialization as an independent step.

\vspace{1mm}
\noindent\textbf{Neural GS Derivation from Shared Features.}  
Unlike the vanilla 3DGS~\cite{kerbl2023-3dgs}, which directly optimizes Gaussian properties, we embed shared features \(\mathbf{f}\) into the GS branch. Specifically, we use a tight MLP \(F_{gs}\) transforms \(\mathbf{f}\) and \(\mathbf{p}\) into Gaussian attributes $\mathcal{A}$ as follows: 
\begin{equation}
\mathcal{A}  = F_{gs}(\mathbf{p}, \mathbf{f}),
\end{equation}
where $\mathcal{A}$ including color SH, opacity $\alpha$, rotation $\mathbf{r}$ and scale $\mathbf{s}$.
Note that the shared information and the newly introduced network are used only during training. The Gaussian attributes can be directly used for inference rendering without compromising the real-time advantage.

\subsection{Residual Vectors in GS branch}
\label{subsec:div}

\noindent\textbf{Residual Feature.}  
The feature at the same spatial point yields distinct information in NeRF and GS branches: NeRF uses the feature to predict density and color, while GS requires an additional derivation of the geometric properties (rotation and scale). Therefore, NeRF-optimized features $\mathbf{f}$ may lack adaptability for predicting Gaussian properties. To address this, we optimize a residual feature vector \(\Delta \mathbf{f}\) for each Gaussian to capture information discrepancies in the shared features.
This refined feature vector maintains consistency with the NeRF branch features while enabling individual Gaussians to fine-tune specific information, thus enhancing the rendering quality of the GS branch.

\vspace{1mm}
\noindent\textbf{Residual Position.}  
Due to potential NeRF fitting errors and differing spatial perception caused by the GS branch, the Gaussian position $\mathbf{p}$ derived from NeRF branch initialization may not fully suit the GS branch. Therefore, in addition to the residual feature, we optimize a residual position \(\Delta \mathbf{p}\) for each Gaussian to capture subtle spatial adjustments. 

After introducing the discrepancy modeling, Gaussian attributes can be derived based on the adjusted position and feature vectors.
\begin{equation}
\mathcal{A}  = F_{gs}(\mathbf{p}+\Delta \mathbf{p}, \mathbf{f}+\Delta \mathbf{f}),
\end{equation}
where $\Delta \mathbf{p}$ and $\Delta \mathbf{f}$ are modeled as trainable parameters as shown in~\cref{fig:pipeline}.

\subsection{Joint Optimization in Dual-branch}
\label{subsec:joint}

\noindent\textbf{GS-Rays.}  
NeRF requires dense sampling and network queries, which preclude rendering an entire image in a single pass like in 3DGS. To synchronize optimization, we propose rendering NeRF using only partial rays in each iteration. We select rays that connect Gaussian positions with the high opacity in the current view frustum space to the camera origin, which we refer to as GS-Rays: $\mathcal{R}_{gs}$. For the $k$-th training view, its  GS-Rays are determined by the corresponding camera origin $\mathbf{o}^k$ and the ray directions $\mathbf{d}_i^k$.
\begin{equation}
\mathcal{R}^k_{gs} = \{\mathbf{o}^k, \mathbf{d}_i^k\}, ~\mathbf{d}_i^k=\mathbf{p}_i^k-\mathbf{o}^k,
\end{equation}
where $\mathbf{p}_i^k$ is visible Gaussian positions with the high opacity in the $k$-th view. This design ensures that the sampling points in the NeRF branch are distributed as closely as possible to the Gaussian spheres, thereby aligning the scene perception and enhancing the effectiveness of the shared information across different branches, as well as providing the necessary data for subsequent joint optimization.

\vspace{1mm}
\noindent\textbf{Growing and Pruning Operation.}  
In the original 3DGS, Gaussian growth occurs via gradient evaluation, which restricts gradient awareness to regions containing Gaussian spheres, potentially overlooking important blank areas. Inspired by Point-NeRF~\cite{xu2022point-nerf}, we leverage NeRF spatial continuity to address this limitation. 
Specifically, we evaluate the opacity at sampling points in the NeRF branch as follows.
\begin{equation}
    \alpha_{nerf} = 1-\exp(-\sigma_i \delta_i).
    \label{eq:a_nerf}
\end{equation}
New Gaussian spheres are then added at points with high opacity and far from existing Gaussian spheres. 
We regulate NeRF-driven growing to achieve an optimal balance between the number of Gaussian spheres and scene representation accuracy, which mitigates the 3DGS limitation of localized growth perception.

For pruning, we adopt the original 3DGS strategy, relying solely on GS branch information without NeRF assistance. This is because the pruning regions already include Gaussian-perceptible areas.

\vspace{1mm}
\noindent\textbf{Loss Design.} 
During joint training, we design loss functions for single-branch optimization and dual-branch collaboration. For the NeRF branch, we use an L1 norm loss \(\mathcal{L}_{\text{nerf}}^\text{rgb}\) for rendered RGB values and an entropy loss \(\mathcal{L}_\text{nerf}^{\text{en}}\)~\cite{kim2022infonerf-en} for density, which promotes a more concentrated distribution pattern by discouraging density dispersion. 
\begin{equation}
\mathcal{L}_{\text{nerf}} = 
\mathcal{L}_{\text{nerf}}^\text{rgb} + 
\lambda_\text{en} \mathcal{L}_{\text{nerf}}^\text{en}.
\end{equation}
For the GS branch, we use an L1 norm loss \(\mathcal{L}_{\text{gs}}^\text{rgb}\) and SSIM loss \(\mathcal{L}_{\text{gs}}^\text{SSIM}\) for rendered images, along with a volume regularization \(\mathcal{L}_{\text{gs}}^\text{vol}\)~\cite{lu2024scaffold} to minimize Gaussian sphere overlap.
\begin{equation}
\mathcal{L}_{\text{gs}} = \mathcal{L}_{\text{gs}}^{rgb} +
\lambda_\text{SSIM} \mathcal{L}_{\text{gs}}^\text{SSIM}+
\lambda_\text{vol} \mathcal{L}_{\text{gs}}^\text{vol}.
\end{equation}
For dual-branch collaborative loss, we use L1 norm  \(\mathcal{L}_{\text{joint}}^\text{rgb}\) to constrain the rendered pixel values along GS-Rays in the NeRF branch with corresponding GS branch rendered pixels. Additionally, we align the Gaussian opacity ($\alpha$ in Eq.~\ref{eq:gs_render}) with corresponding NeRF opacities ($\alpha_{nerf}$ in Eq.~\ref{eq:a_nerf}) by L1 norm loss \(\mathcal{L}_{\text{joint}}^\text{op}\). Residual features and position residuals are further constrained by L2 norm regularization, denoted as \(\mathcal{L}_{\text{reg}}^\text{fea}\) and \(\mathcal{L}_{\text{reg}}^\text{pos}\), to encourage NeRF and GS branches to learn common spatial properties while providing mutual regularization against overfitting. The overall loss function during joint optimization is as follows:
\begin{align}
\mathcal{L}_{\text{total}} = & \mathcal{L}_\text{gs} +
\lambda_\text{nerf} \mathcal{L}_{\text{nerf}} + 
\lambda_\text{rgb} \mathcal{L}_{\text{joint}}^\text{rgb} + \nonumber \\
& \lambda_\text{op} \mathcal{L}_{\text{joint}}^\text{op} +
\lambda_\text{fea} \mathcal{L}_{\text{reg}}^\text{fea} +
\lambda_\text{pos} \mathcal{L}_{\text{reg}}^\text{pos}.
\end{align}

\section{Experiments}
\label{sec:exp}

\begin{table*}[t]
\centering
\caption{\textbf{Quantitative comparison on real-world datasets.} Colors denote the \colorbox[rgb]{1, 0.7, 0.7}{1st}, \colorbox[rgb]{1, 0.85, 0.7}{2nd}, and \colorbox[rgb]{1, 1, 0.7}{3rd} best-performing model.
}
\label{tab:full_real_scn}
\resizebox{0.75\linewidth}{!}{

\begin{tabular}{c|ccc|ccc|ccc}
\toprule
\multirow{2}[4]{*}{Method} & \multicolumn{3}{c|}{DeepBlending} & \multicolumn{3}{c|}{Mip-NeRF360} & \multicolumn{3}{c}{Tanks\&Temples		} \\
\cmidrule{2-10}      & PSNR \(\uparrow\)  & SSIM \(\uparrow\)  & LPIPS \(\downarrow\) & PSNR \(\uparrow\)  & SSIM \(\uparrow\)  & LPIPS \(\downarrow\) & PSNR \(\uparrow\)  & SSIM \(\uparrow\)  & LPIPS \(\downarrow\) \\
\midrule
INGP~\cite{mueller2022instant} & 23.62 & 0.797 & 0.423 & 26.43 & 0.725 & 0.339 & 21.72 & 0.723 & 0.330 \\[1pt]
3DGS~\cite{kerbl2023-3dgs} & 29.42  & 0.899 & \cellcolor{tabthird}0.247  & 27.49 & \cellcolor{tabsecond}0.813 & \cellcolor{tabthird}0.222 & 23.69 & 0.844 & 0.178 \\[1pt]
C3DGS~\cite{niedermayr2024-c3dgs} & 29.79  & 0.901  & 0.258  & 27.08 & 0.798 & 0.247 & 23.32 & 0.831 & 0.201 \\[1pt]
Scaffold-GS~\cite{lu2024scaffold} & \cellcolor{tabsecond}30.21 & \cellcolor{tabsecond}0.906  & 0.254  & 27.5  & 0.806 & 0.252 & 23.96 & \cellcolor{tabsecond}0.853 & \cellcolor{tabthird}0.177 \\[1pt]
Hash-GS~\cite{chen2025hac} & \cellcolor{tabthird}29.98  & \cellcolor{tabthird}0.902  & 0.269 & \cellcolor{tabthird}27.53 & \cellcolor{tabthird}0.807 & 0.238 & \cellcolor{tabsecond}24.04 & 0.846 & 0.187 \\[1pt]
VDGS~\cite{malarz2023-vdgs} & 29.54 & \cellcolor{tabsecond}0.906  & \cellcolor{tabsecond}0.243 & \cellcolor{tabsecond}27.64 & \cellcolor{tabsecond}0.813 & \cellcolor{tabsecond}0.220 & \cellcolor{tabthird}24.02 & \cellcolor{tabthird}0.851 & \cellcolor{tabsecond}0.176 \\[1pt]
\textbf{Ours} & \cellcolor{tabfirst}30.70 & \cellcolor{tabfirst}0.912 & \cellcolor{tabfirst}0.237 & \cellcolor{tabfirst}28.32 & \cellcolor{tabfirst}0.817 & \cellcolor{tabfirst}0.210 & \cellcolor{tabfirst}24.44 & \cellcolor{tabfirst}0.860 & \cellcolor{tabfirst}0.161 \\
\bottomrule
\end{tabular}%

}
\end{table*}

\begin{figure*}[t]
  \centering
   \includegraphics[width=0.9\linewidth]{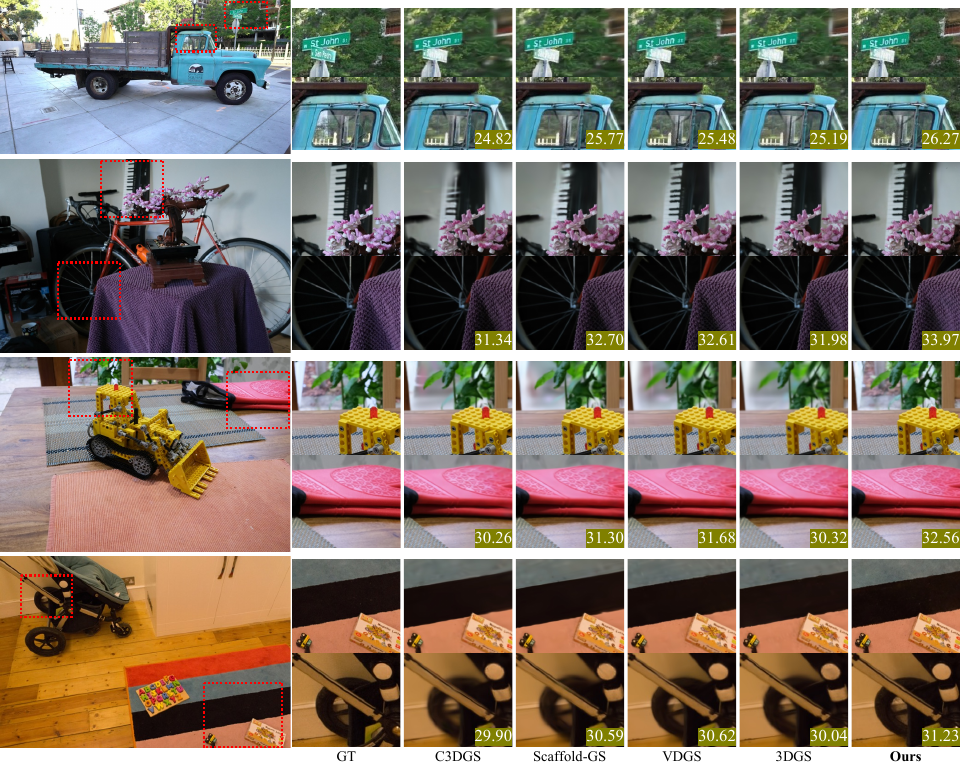}
   \caption{\textbf{Qualitative comparison on real-world datasets}. The numbers indicate the PSNR. Our method demonstrates a significant advantage over 3DGS and its variants, achieving a more faithful representation of scene details.
   }
   \label{fig:compare_full_view}
\end{figure*}

\begin{figure*}[t]
  \centering
   \includegraphics[width=1.\linewidth]{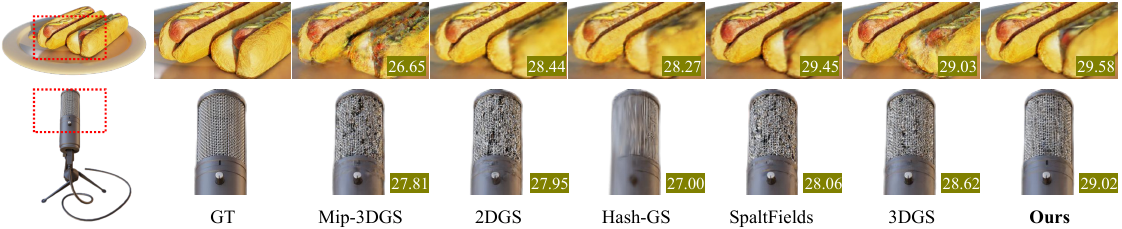}
   \caption{\textbf{Qualitative comparison under 12 input views on the Blender dataset}. The numbers indicate the PSNR.
   }
   \label{fig:sparse}
\end{figure*}

\subsection{Implementation Details}
\noindent\textbf{Training and Optimization Details}.
Following the original 3DGS method configurations, we implement \methodname{} in PyTorch. For the Hash network, we set 16 different level grids, each outputting 2-dimensional features, yielding a 32-dimensional feature vector for a spatial point.
During the NeRF branch pre-training, each batch contains 8,192 rays and is trained for 10 epochs. For real-world datasets, we initialize using 1,000,000 points sampled at an 8:2 ratio from edge rays and random rays, while Blender datasets are initialized with 100,000 points. To enhance efficiency during joint training, the NeRF branch renders 4,096 rays sampled from GS-Rays.
We set \(\lambda_{\text{en}}\), \(\lambda_{\text{SSIM}}\), and \(\lambda_{\text{vol}}\) to 1e-4, 0.2, and 1e-3, respectively, while \(\lambda_{\text{nerf}}\), \(\lambda_{\text{rgb}}\), \(\lambda_{\text{op}}\), \(\lambda_{\text{fea}}\), and \(\lambda_{\text{pos}}\) are set to 0.1, 0.05, 1e-3, 1e-4, and 1e-4 respectively. 
NeRF-assisted Gaussian growth occurs every 100 iterations, with a maximum of 200 new additions. Joint training iterates 30k for full-view datasets and 8k for sparse-view scenes. All experiments are conducted on an NVIDIA A100 GPU.

\vspace{1mm}
\noindent\textbf{Datasets and Metrics}.
We report experimental results on real-world datasets, including Mip-NeRF360 (all 9 scenes)~\cite{barron2022mipnerf-360}, Tanks\&Temples~\cite{knapitsch2017tanksandtemples} DeepBlending~\cite{hedman2018deepblending}, and the Blender dataset~\cite{mildenhall2020nerf}. Evaluation metrics include PSNR, SSIM~\cite{wang2004image-ssim}, and LPIPS~\cite{zhang2018-lpips}. Additionally, we compare metrics for training time (minutes), storage size (MB), and rendering speed (FPS) to assess the model’s compactness and efficiency.

\vspace{1mm}
\noindent\textbf{Baselines}. Our method is focused on enhancing GS branch performance, so we primarily compare it with 3DGS~\cite{kerbl2023-3dgs} and its variants, including C3DGS~\cite{niedermayr2024-c3dgs}, Scaffold-GS~\cite{lu2024scaffold}, Mip3DGS~\cite{yu2024mip-gs}, and 2DGS~\cite{huang20242dgs}. We also compare methods incorporating NeRF properties such as Hash-GS~\cite{chen2025hac},
and VDGS~\cite{malarz2023-vdgs}, as well as SplatFields~\cite{mihajlovic2025splatfields}, which is specifically designed for sparse-view scenes.

\subsection{Comparison}

\begin{table}[htb]
\centering
\caption{\textbf{Quantitative comparison of using different numbers of input views on Blender dataset}. Our \methodname{} maintains high performance when the scene input views are reduced.
}
\label{tab:sparse}
\resizebox{1.\linewidth}{!}{
\begin{tabular}{c|cc|cc|cc}
\toprule
\multirow{2}[4]{*}{Method} & \multicolumn{2}{c|}{Full views} & \multicolumn{2}{c|}{12 views} & \multicolumn{2}{c}{8 views} \\
\cmidrule{2-7}      & PSNR\(\uparrow\)  & SSIM\(\uparrow\)  & PSNR\(\uparrow\)  & SSIM\(\uparrow\)  & PSNR\(\uparrow\)  & SSIM\(\uparrow\) \\
\midrule
SparseNeRF~\cite{wang2023sparsenerf} & 32.46 & 0.957 & 22.92 & 0.875 & 22.20 & 0.861 \\[3pt]
INGP~\cite{mueller2022instant} & 33.18 & 0.960 & 22.68 & 0.875 & 21.87 & 0.860 \\[3pt]
3DGS~\cite{kerbl2023-3dgs} & 33.32 & \cellcolor{tabfirst}0.970 & 25.29 & 0.900 & 22.93 & 0.866 \\[3pt]
Mip3DGS~\cite{yu2024mip-gs} & 33.36 & \cellcolor{tabsecond}0.969 & 24.86 & 0.898 & 22.37 & 0.862 \\[3pt]
Scaffold-GS~\cite{lu2024scaffold} & \cellcolor{tabsecond}33.41 & 0.966 & 23.82 & 0.874 & 21.53 & 0.836 \\[3pt]
2DGS~\cite{huang20242dgs} & 33.07 & 0.964 & \cellcolor{tabthird}25.62 & \cellcolor{tabsecond}0.911 & 23.04 & 0.877 \\[3pt]
Hash-GS~\cite{chen2025hac} & 33.24 & \cellcolor{tabthird}0.967 & 25.36 & \cellcolor{tabthird}0.909 & \cellcolor{tabthird}23.14 & \cellcolor{tabthird}0.879 \\[3pt]
VDGS~\cite{malarz2023-vdgs} & \cellcolor{tabthird}33.37 & \cellcolor{tabsecond}0.969 & 24.77 & 0.898 & 22.88 & 0.872 \\[3pt]
SplatFields~\cite{mihajlovic2025splatfields} & 33.25 & 0.966 & \cellcolor{tabsecond}25.80 & \cellcolor{tabsecond}0.911 & \cellcolor{tabfirst}23.98 & \cellcolor{tabfirst}0.889 \\[3pt]
\textbf{Ours} & \cellcolor{tabfirst}33.71 & \cellcolor{tabfirst}0.970 & \cellcolor{tabfirst}26.34 & \cellcolor{tabfirst}0.912 & \cellcolor{tabsecond}23.92 & \cellcolor{tabsecond}0.881 \\
\bottomrule
\end{tabular}%

}
\end{table}

We conduct extensive quantitative and qualitative comparisons with state-of-the-art methods on both full and sparse datasets. 
As our primary focus is on the impact of NeRF integration on GS performance, all results, unless otherwise noted, use the GS branch as the final output of \methodname{}.

\vspace{1mm}
\noindent\textbf{Full View Scene.}
We optimize \methodname{} using the default full training data on multiple benchmark datasets. 
Comparative results are shown in Table~\ref{tab:full_real_scn}, where our approach significantly outperforms the vanilla 3DGS model and other state-of-the-art methods across PSNR, SSIM, and LPIPS metrics.
Qualitative experiments in~\cref{fig:compare_full_view} demonstrate our method's superior capability in capturing high-frequency textures and fine geometric details while better reflecting lighting conditions.
Notably, compared to other methods that incorporate NeRF-like concepts, such as VDGS and Hash-GS, 
\methodname{} achieves even more substantial improvements. This indicates that our dual-branch joint optimization framework is more effective than simple NeRF initialization or directly adapting NeRF implicit concepts, validating \methodname{} as a robust framework for integrating diverse 3D representation approaches.

\begin{figure}[t]
  \centering
   \includegraphics[width=1\linewidth]{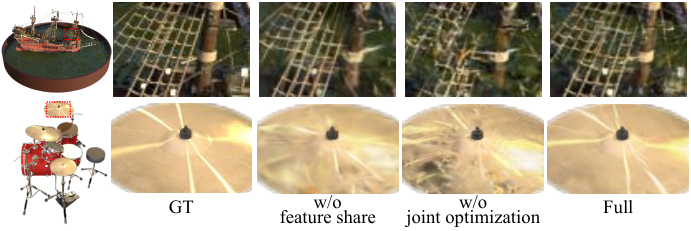}
   \caption{\textbf{Impact of feature share and joint optimization on sparse view scenes}. These two key designs enable mutual regularization constraints between NeRF and GS branches, significantly improving the visual quality of \methodname{} in sparse views.
   }
   \label{fig:ablation-drums}
\end{figure}

\begin{figure}[ht]
  \centering
   \includegraphics[width=0.95\linewidth]{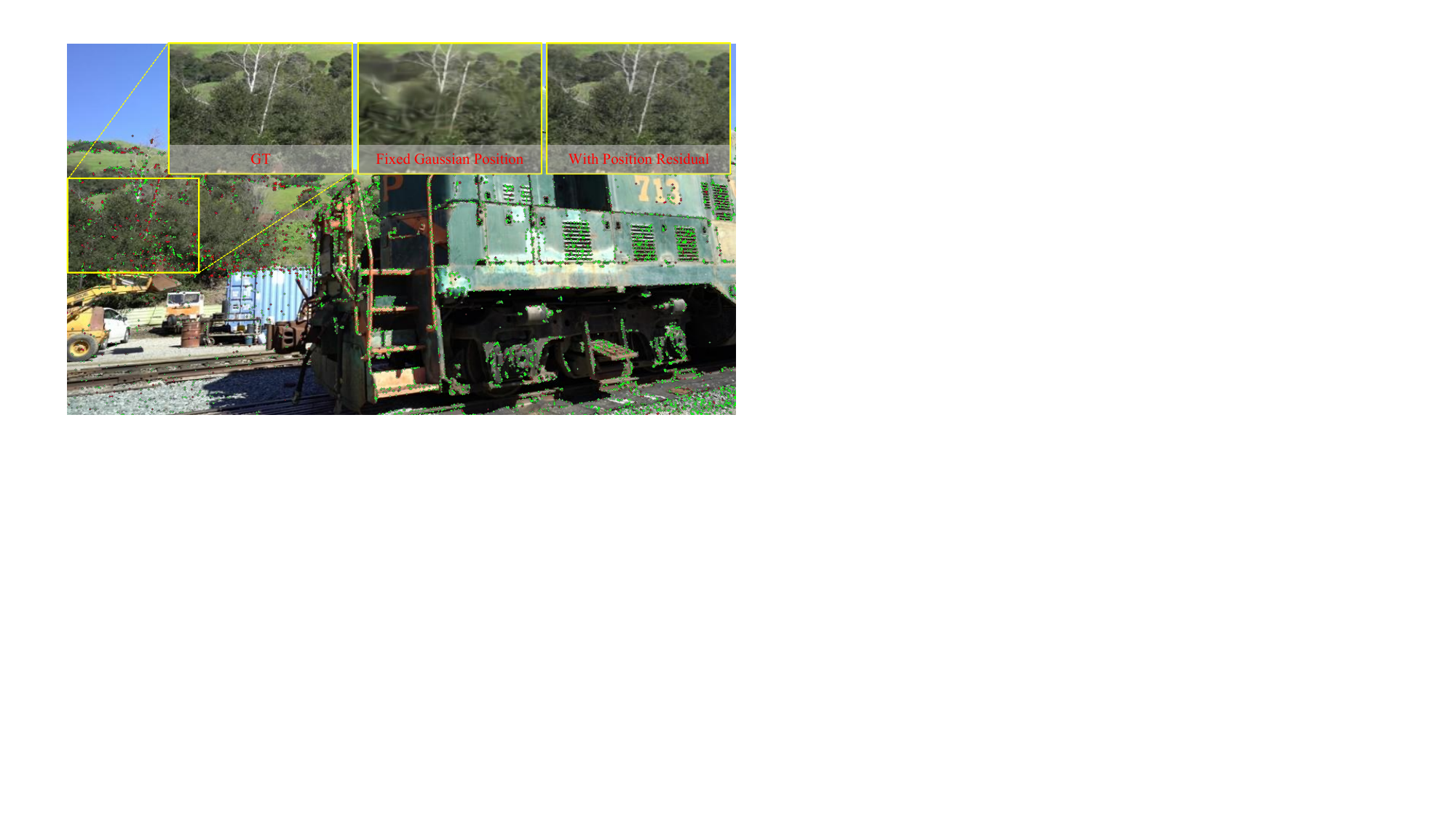}
   \caption{\textbf{Visualization of position residuals}. The points represent the initial Gaussian positions, with the top 20\% of points having the largest optimized residuals highlighted in red. We compare this with the results obtained by fixed Gaussian positions during training, demonstrating the importance of the residual vectors for personalized adaptation in the GS branch.
   }
   \vspace{-3mm}
   \label{fig:vis-div}
\end{figure}

\begin{table}[t]
\centering
\caption{\textbf{Comparison of model efficiency with 3DGS}. 
We report the FPS, model size (MB), training time (minutes) and PSNR. The $\text{3DGS}_L$ denotes longer iterative training (50k) for 3DGS.
}
\label{tab:efficiency}
\resizebox{0.95\linewidth}{!}{

\begin{tabular}{c|c@{\hspace{3mm}}c@{\hspace{3mm}}c@{\hspace{3mm}}c|c@{\hspace{3mm}}c@{\hspace{3mm}}c@{\hspace{3mm}}c}
\toprule
\multirow{2}[4]{*}{Method} & \multicolumn{4}{c|}{DeepBlending} & \multicolumn{4}{c}{Mip-NeRF360} \\
\cmidrule{2-9}      & FPS\(\uparrow\)   & Size\(\downarrow\)  & Time\(\downarrow\)  & PSNR\(\uparrow\)  & FPS\(\uparrow\)   & Size\(\downarrow\)  & Time\(\downarrow\)  & PSNR\(\uparrow\) \\
\midrule
3DGS  & 105   & 672   & \textbf{36.1}  & 29.42 & 101   & 729   & \textbf{41.5}  & 27.49 \\[3pt]
$\text{3DGS}_L$ & 104   & 678   & 55.7  & 29.50 & 102   & 733   & 67.9  & 27.57 \\[3pt]
\textbf{Ours} & \textbf{122}   & \textbf{526}   & 51.7  & \textbf{30.70} & \textbf{134 }  & \textbf{564}   & 60.3  & \textbf{28.32} \\
\bottomrule
\end{tabular}
}
\vspace{-3mm}
\end{table}

\vspace{1mm}
\noindent\textbf{Sparse View Scene.}
Through shared spatial positions and corresponding encoded features, different branches within \methodname{} can more comprehensively perceive and learn from limited 3D scene information. Additionally, the collaborative optimization between NeRF and GS branches, facilitated by this shared information, creates mutual constraints and regularization effects, mitigating overfitting, which is crucial for modeling scenes under sparse views. To validate this, we perform sparse-view comparisons with baseline methods, as shown in Table~\ref{tab:sparse}.
Across various sparsity levels, \methodname{} consistently surpasses corresponding baselines. Remarkably, \methodname{} achieves performance comparable to or even surpassing the SplatField method, which is specifically designed for sparse-view settings.
\cref{fig:sparse} also provides a qualitative illustration of these improvements. Furthermore, it can be observed that the performance gap between \methodname{} and baseline methods in sparse views is more pronounced than in full views, affirming the effectiveness of our method’s regularization. We further conduct relevant experiments in the next subsection.

\subsection{Qualitative Analysis of \methodname{}}

\noindent\textbf{Regularization Effect}.
By introducing the spatial continuity of NeRF, \methodname{} establishes self-correlation across different Gaussian spheres in 3DGS. Additionally, feature sharing, cross-branch loss constraints, and joint optimization enable mutual regularization between NeRF and 3DGS. In~\cref{fig:ablation-drums}, we illustrate the critical role of NeRF-GS in preventing overfitting to the scene.  When associations between two branches are directly removed, such as feature sharing, loss constraints during joint training, etc., the \methodname{} shows large visual quality degradation. 

\vspace{1mm}
\noindent\textbf{Visualization of Discrepancy}.
Errors introduced during NeRF pre-training and inherent disparities between NeRF and 3DGS can impede the GS branch's ability to effectively model a 3D scene from NeRF-shared information. \methodname{} addresses this challenge through a residual mechanism. An example is shown in~\cref{fig:vis-div}, incorporating positional residuals allows the GS branch to adjust Gaussian positions, avoiding artifacts that could arise from only adjusting Gaussian shapes under fixed positions.

\vspace{1mm}
\noindent\textbf{Model Efficiency.}
While \methodname{} bridges two distinct 3D representation models, it maintains their independence. Post-joint training, each branch can retain only its effective components, preserving the original single-branch inference speed. 
As shown in Table~\ref{tab:efficiency}, our method maintains GS real-time rendering capabilities while requiring less storage than the original 3DGS approach. This is because our initialization and Gaussian growing strategies reduce Gaussian spheres. For example, on the DeepBlending dataset, vanilla 3DGS uses 2,461,023 Gaussians, while ours uses only 1,926,336.
We also compare it with an extended-training version of $\text{3DGS}_L$, showing \methodname{} outperforms 3DGS even with similar training time. This suggests that integrating the NeRF branch is a worthwhile trade-off despite the increase in training time.

\begin{table}[t]
\centering
\caption{\textbf{Ablation of different components in \methodname{} on Tank\&Temples and DeepBlending datasets}. Numbers represent the PSNR metric. See Section~\ref{sec:sub-ablation} for discussion.}
\label{tab:ablation}
\resizebox{1.\linewidth}{!}{
\begin{tabular}{c@{\hspace{2mm}}|c@{\hspace{2mm}}c@{\hspace{2mm}}c@{\hspace{1mm}}|c@{\hspace{1mm}}c@{\hspace{1mm}}c}
\toprule
\multirow{2}[4]{*}{} & \multicolumn{3}{c|}{Tanks\&Temples} & \multicolumn{3}{c}{DeepBlending} \\
\cmidrule{2-7}      & Truck & Train & Avg   & Drjohnson & Playroom & Avg \\
\midrule
      & \multicolumn{6}{c}{Ablation of  sharing mechanisms} \\
\cmidrule{2-7}w/o Edge-based Init & 24.06 & 21.17 & 22.61 & 28.65 & 29.8  & 29.8 \\[1pt]
w/o Feature Share & 25.74 & 22.12 & 23.93 & 29.54 & 30.91 & 30.22 \\
\midrule
      & \multicolumn{6}{c}{Ablation of residual vectors} \\
\cmidrule{2-7}w/o Residual Feature & 25.88 & 22.31 & 24.09 & 29.76 & 30.84 & 30.3 \\[1pt]
w/o Residual Position & 25.97 & 22.35 & 24.16 & 29.89 & 31.01 & 30.45 \\
\midrule
      & \multicolumn{6}{c}{Ablation of joint optimization } \\
\cmidrule{2-7}w/o $\mathcal{L}_{\text{joint}}^\text{fea}$ & 26.16 & 22.51 & 24.33 & 30.05 & 30.88 & 30.46 \\[6pt]
w/o $\mathcal{L}_{\text{joint}}^\text{pos}$ & 26.09 & 22.46 & 24.27 & 30.1  & 31.03 & 30.56 \\[6pt]
w/o $\mathcal{L}_{\text{joint}}^\text{op}$ & 26.3  & 22.4  & 24.35 & 30.02 & 31.17 & 30.60 \\[6pt]
w/o $\mathcal{L}_{\text{joint}}^\text{rgb}$ & 26.14 & 22.48 & 24.31 & 30.21 & 30.97 & 30.59 \\[6pt]
w/o GS-Rays & 25.82 & 22.26 & 24.04 & 29.85 & 30.6  & 30.22 \\
\midrule
\textbf{Full} & 26.27 & 22.61 & \textbf{24.44} & 30.17 & 31.23 & \textbf{30.7} \\
\bottomrule
\end{tabular}%

}
\end{table}

\subsection{Ablation Studies}
\label{sec:sub-ablation}

\noindent\textbf{Impact of Sharing Mechanisms.}
In \methodname{}, information exchange manifests through spatial co-utilization and feature sharing. We propose a scene-edge-based initialization scheme as in Eq.~\ref{eq:init} and compare it with the alternative initialization from SFM, denoted as `w/o Edge-based Init'.
Moreover, to examine the effect of feature sharing, we directly train the GS branch with learnable feature parameters, remarked as `w/o Feature Share'. The ablation results in Table~\ref{tab:ablation} indicate that our proposed initialization significantly outperforms the alternatives. Likewise, the feature-sharing scheme across NeRF and GS exhibits irreplaceable positive impacts on the full scene.

\vspace{1mm}
\noindent\textbf{Impact of Residual Vectors.}
As discussed in Sec~\ref{subsec:div}, the GS branch needs to derive different geometric information from NeRF's shared features and initial points, suggesting the need for differentiated information encoding. 
To achieve this, we introduce residual strategies for both features and Gaussian positions. Removing these terms results in significant performance degradation, as shown in Table~\ref{tab:ablation}. 
This indicates that, in addition to information sharing, enabling each branch to learn adapted and differentiated information is also critical. In contrast, previous methods such as Scaffold-GS, Hash-GS and VDGS that merely incorporate NeRF characteristics overlooked this distinction, thereby offering limited performance improvement.

\vspace{1mm}
\noindent\textbf{Impact of Joint Optimization Strategy.}
Our joint optimization process incorporates several key components to ensure efficient and effective training between the NeRF and GS branches. The GS-Rays strategy directs NeRF to focus on areas that are essential for the GS branch during rendering, effectively enabling efficient information exchange and mutual enhancement between branches.  The term `w/o GS-Rays' in Table~\ref{tab:ablation} shows performance decline when replacing GS-Rays with an equal number of random rays.
We also evaluate the effectiveness of newly introduced loss terms, as indicated in Table~\ref{tab:ablation}. Removing mutual constraints between branch outputs leads to performance degradation. Furthermore, applying regularization constraints to shared information to prevent excessive branch discrepancy enhances model performance.

\section{Limitations}
\label{sec:limitation}
Although \methodname{} fundamentally turns NeRF and 3DGS from competitors into collaborators and achieves superior performance, 
it also increases method complexity and computational overhead. Certain components within the two full pipelines may be redundant when combined. Developing a more compact and streamlined integration strategy could enhance our framework’s applicability and improve its interpretability in future research.

\section{Conclusion}
\label{sec:conclusion}
In this study, we introduce \methodname{}, a novel framework that combines implicit neural radiance fields with Gaussian splatting. Its core innovation lies in dual-branch collaborative design, comprising three key components: shared information in positional spaces and features, residual vectors to model inherent inter-branch differences, and joint optimization via GS-Rays alignment of intermediate results and rendering outputs, as well as adaptive Gaussian controls. 
These strategies effectively address several limitations of 3DGS, including initialization dependency, limited spatial awareness, insufficient Gaussian sphere correlation, and overfitting in sparse-view scenes. 
Experimental results demonstrate that \methodname{} achieves state-of-the-art performance, offering new insights into the fusion of NeRF and 3DGS as an efficient hybrid approach for 3D scene representation.

\section*{Acknowledgments}
This work is supported by the National Natural Science Foundation of China (U24B6013) and China Scholarship Council (202406020139).

{
    \small
    \bibliographystyle{ieeenat_fullname}
    \bibliography{main}
}

\clearpage
\setcounter{page}{1}

\twocolumn[{%
	\renewcommand
	\twocolumn[1][]{#1}%
	\maketitlesupplementary
	\begin{center}
		\centering
            \includegraphics[width=1.\textwidth]{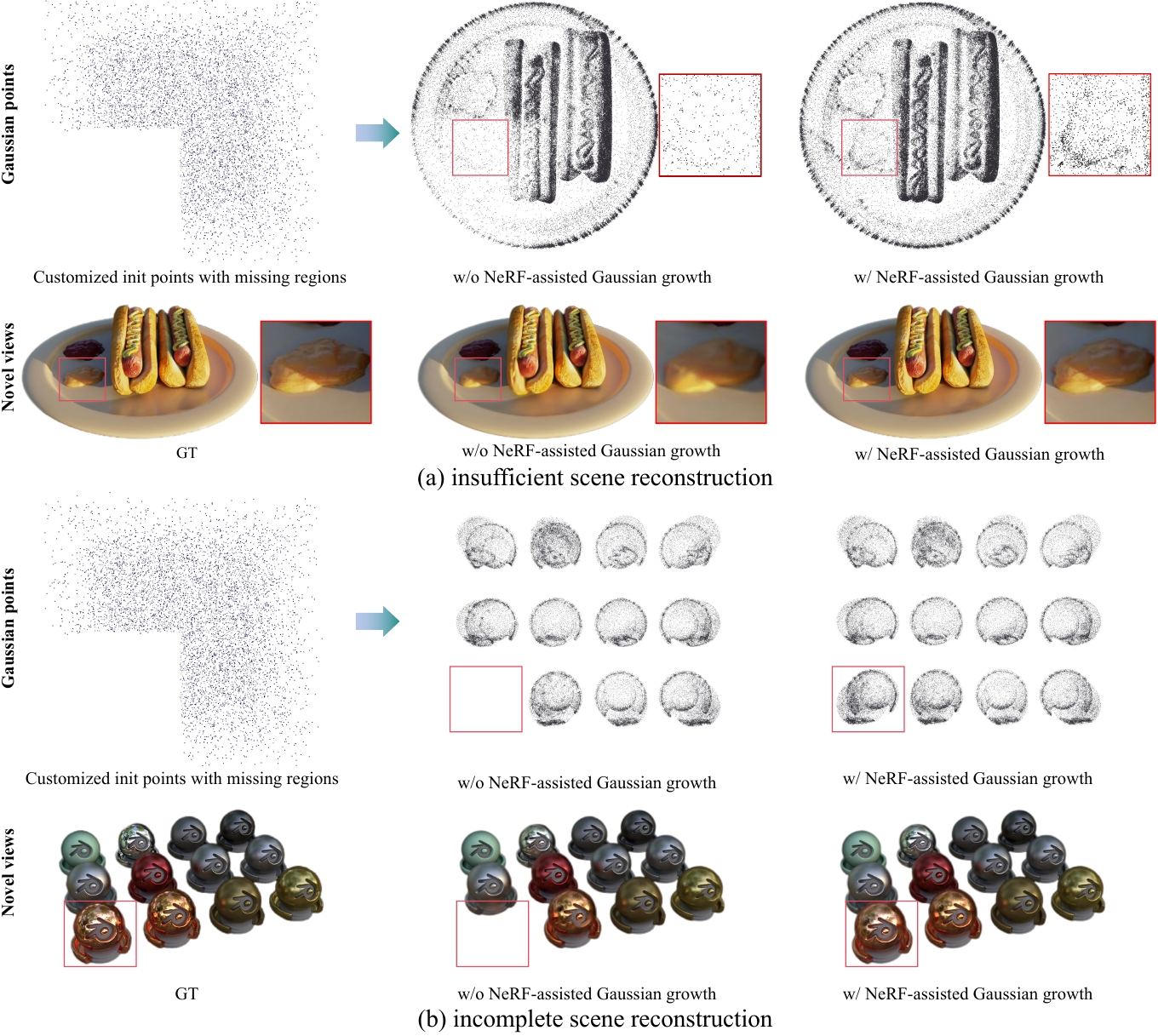}
		\captionof{figure}{
            \textbf{Impact of NeRF-assisted Gaussian growth}. 
            We initialize 3DGS using point clouds with missing regions to evaluate its scene perception range and sensitivity to initialization. Without NeRF-assisted Gaussian growth, 3DGS exhibits insufficient reconstruction (a) or incomplete reconstruction (b) in the missing areas. However, when employing the proposed NeRF-assisted Gaussian growth strategy in our method, these missing regions are successfully reconstructed. This demonstrates that NeRF significantly enhances the perception range of 3DGS, reducing its sensitivity to initialization and improving visual quality.
		}
		\label{fig:nerf-growth}
	\end{center}
}]

\begin{figure*}[t]
  \centering
   \includegraphics[width=1\linewidth]{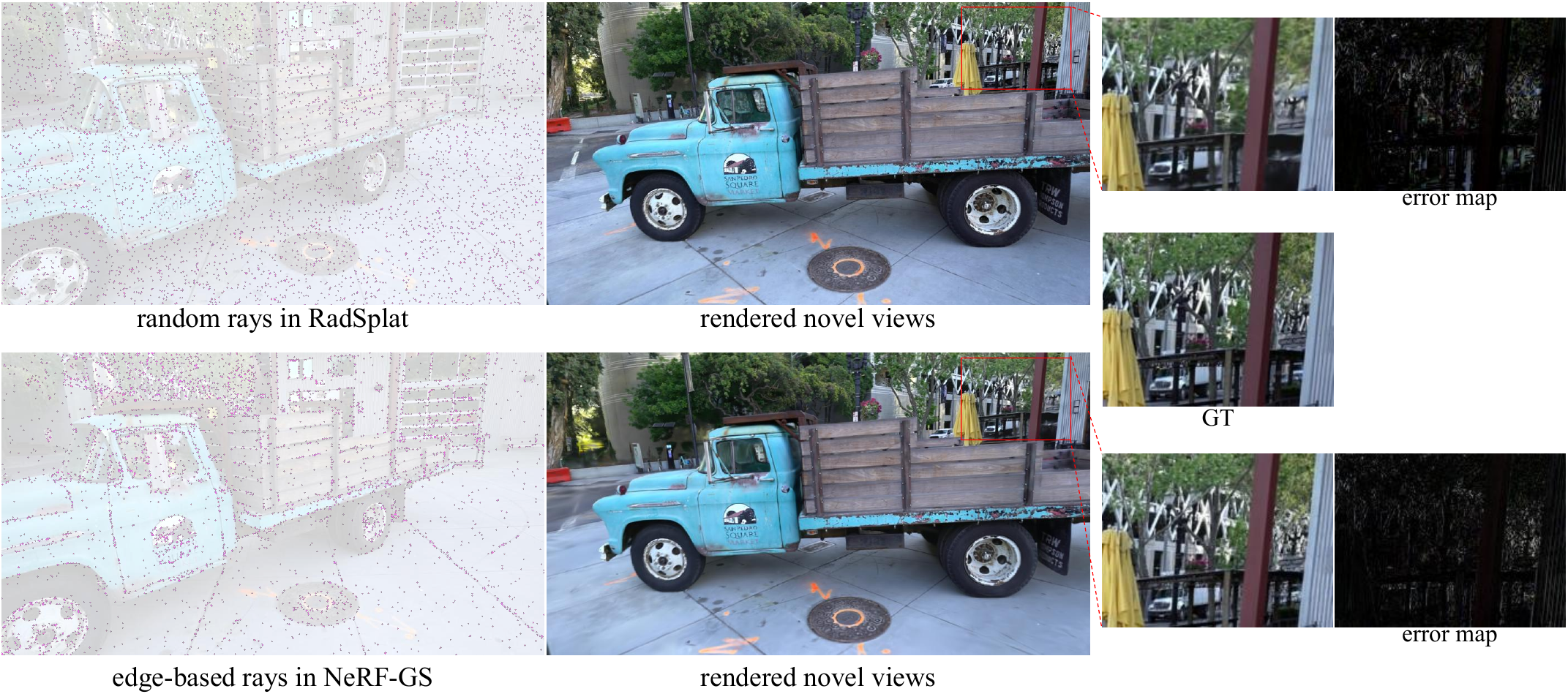}
    \caption{ \textbf{Comparison of initialization with RadSplat}. \methodname{}  focuses more on the contours of the scene during ray sampling, alleviating the burden of position optimization in the GS branch while achieving superior visual results in regions with complex textures.}
   \label{fig:edge-init}
\end{figure*}

\begin{figure*}[t]
  \vspace{3mm}
  \centering
   \includegraphics[width=1\linewidth]{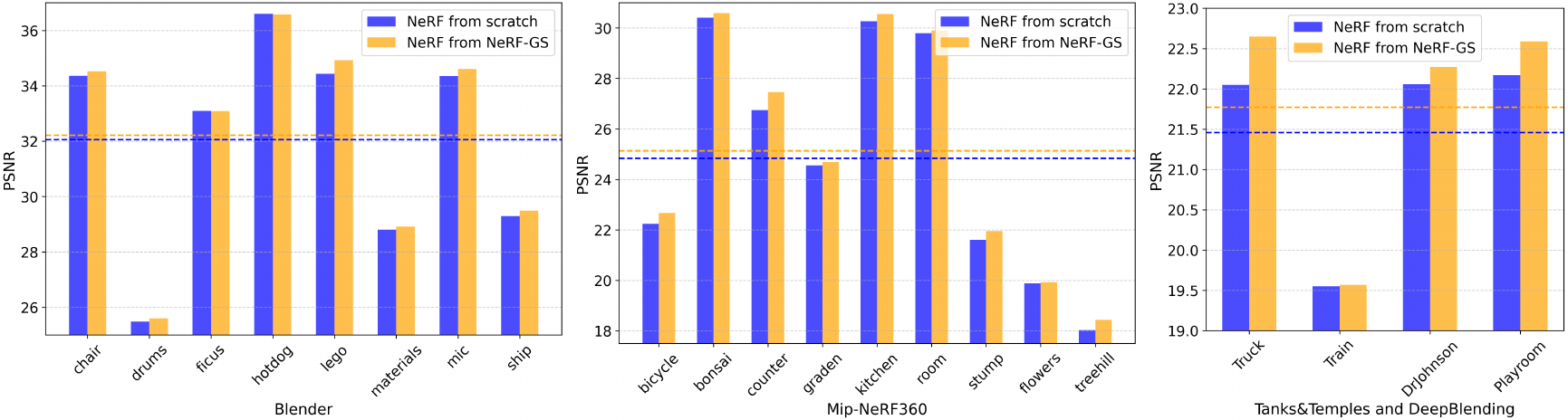}
   \caption{\textbf{Impact of joint optimization on the NeRF branch}. The dashed line indicates the mean PSNR. Given equivalent training iterations, the NeRF obtained through NeRF-GS outperforms training this NeRF independently. This demonstrates that dual-branch training not only benefits the GS branch but also enhances the performance of the NeRF branch.
   }
   \label{fig:nerf-branch}
\end{figure*}

\section{Analysis of Gaussian Adaptive Control from NeRF Branch}
The continuous spatial representation of NeRF enables queries at any spatial location, allowing it to perceive the entire 3D scene. In contrast, individual Gaussian sphere in 3DGS has a limited perceptual range, making 3DGS sensitive to initialization and less effective in adaptive control. 
As shown in Figure~\ref{fig:nerf-growth}, we deliberately design a Gaussian initialization with missing regions in certain spatial areas. 
After iterative optimization, it can be observed that GS allocates a limited number of Gaussians to these regions without assistance from the NeRF branch, and in extreme cases, it fails to perceive the missing areas entirely, resulting in poor or incomplete scene reconstruction. Conversely, our NeRF-assisted adaptive control strategy successfully senses these regions, significantly enhancing the global perceptual capability of the GS branch and reducing its sensitivity to initialization.

\section{Analysis of Edge-based Initialization}
\methodname{} utilizes pre-trained NeRF to obtain candidate Gaussian positions. To enhance initialization efficiency, we incorporate an edge detection step that pre-identifies critical rays and increases their sampling probability during initialization. This design is predicated on the observation that Gaussian spatial distribution should ideally align with the contours of the actual 3D scene, with more Gaussians in textured areas and fewer in blank areas. 
In the baseline RadSplat, rays are sampled uniformly at random without discrimination,  which we consider inefficient. 
To illustrate this, we conduct a visualization experiment in Figure~\ref{fig:edge-init}, showing that our approach yields a Gaussian distribution that clusters around areas rich in texture, with fewer Gaussians in low-texture or empty regions. The rendering results demonstrate that our edge-based initialization method effectively captures complex scene textures, outperforming uniform sampling in accurately representing the scene.

\section{Analysis of NeRF Branch Performance}
\noindent \textbf{Mutual Promotion between NeRF and GS Branches}. 
While the primary aim of this work is to leverage NeRF characteristics to address 3DGS limitations, we have found that the GS branch also positively impacts the NeRF branch during joint training. 
As depicted in Figure~\ref{fig:nerf-branch}, the NeRF branch trained jointly with the GS branch outperforms an independently optimized NeRF under the same number of iterations. 
This improvement arises from feature sharing and joint loss constraints between NeRF and GS branches, which enhance NeRF optimization as well. The simultaneous performance gains of both branches further confirm the complementary relationship between NeRF and 3DGS, offering insights for exploring integration with other forms of 3D representation.

\noindent \textbf{Compare with Structurally Similar NeRF Method}. 
We further compare the NeRF branch to the GS branch and the Instant-NGP~\cite{mueller2022instant} based on the same hash structure. It should be noted that this article focuses more on the improvement of the GS branch performance by NeRF, where we observe a significant performance improvement in the GS branch.

\begin{table}[t]
  \centering
  \caption{\textbf{Additional comparisons}. We evaluate the performance of the NeRF branch in NeRF-GS and compare it to Instant-NGP~\cite{mueller2022instant}, which also utilizes a hash-based structure.}

\resizebox{1.\linewidth}{!}{
        \begin{tabular}{c|ccc|ccc|ccc}
        \toprule
        \multirow{2}[4]{*}{Method} & \multicolumn{3}{c|}{DeepBlending} & \multicolumn{3}{c|}{Tanks\&Temples		} & \multicolumn{3}{c}{Mip-NeRF360} \\
        \cmidrule{2-10}      & PSNR\(\uparrow\) & SSIM\(\uparrow\) & LPIPS\(\downarrow\) & PSNR\(\uparrow\) & SSIM\(\uparrow\) & LPIPS\(\downarrow\) & PSNR\(\uparrow\) & SSIM\(\uparrow\) & LPIPS\(\downarrow\) \\
        \midrule
        Instant-NGP & 23.62 & 0.797 & 0.423 & 21.72 & 0.723 & 0.330 & 26.43 & 0.725 & 0.339 \\
        $\text{Branch}_\text{NeRF}$ & 22.43 & 0.784 & 0.441 & 21.11 & 0.718 & 0.338 & 25.12 & 0.722 & 0.343 \\
        $\text{Branch}_\text{GS}$ & 30.70 & 0.910 & 0.245 & 24.44 & 0.860 & 0.172 & 28.32 & 0.824 & 0.217 \\
        \bottomrule
        \end{tabular}%
    }
  \label{tab:rebuttal-compare}%
\end{table}%

\begin{table}[tbp]
  \centering
  \caption{\textbf{Additional ablation studies}. Numbers are PSNR metric. $\text{D}_\text{3dgs}^{\text{random}}$ and $\text{D}_\text{3dgs}^{\text{edge}}$ denote direct optimizing 3DGS after initialization using the random initialization and the proposed edge-based initialization, respectively.}
  \resizebox{0.9\linewidth}{!}{
\begin{tabular}{c|ccc|ccc}
\toprule
\multirow{2}[4]{*}{} & \multicolumn{3}{c|}{Tanks\&Temples} & \multicolumn{3}{c}{DeepBlending} \\
\cmidrule{2-7}      & Truck & Train & Avg   & Drjohnson & Playroom & Avg \\
\midrule
w/o $\mathcal{L}_{\text{gs}}^\text{vol}$ & 26.10  & 22.48  & 24.29  & 30.02  & 31.07  & 30.55  \\
w/o $\mathcal{L}_{\text{nerf}}$ & 25.44  & 21.15  & 23.30  & 28.79  & 29.46  & 29.13  \\
\midrule
\(\text{D}_\text{3dgs}^\text{random}\) & 25.46  & 21.83  & 23.65  & 29.07  & 29.92  & 29.50  \\[3pt]
\(\text{D}_\text{3dgs}^\text{edge}\) & 25.87  & 22.11  & 23.99  & 29.40  & 30.38  & 29.89  \\
\midrule
\textbf{Full} & 26.27  & 22.61  & 24.44  & 30.17  & 31.23  & 30.70  \\
\bottomrule
\end{tabular}%

    }
  \label{tab:rebuttal-ablation}%
\end{table}%
\section{Additional Ablation Studies}
We further conduct ablation studies on additional loss terms, including the introduced volume regularization~\cite{lu2024scaffold} and the overall loss term of the NeRF branch, \( \mathcal{L}_{\text{nerf}} \). Additionally, we evaluate the performance of directly optimizing 3DGS after initialization using the random initialization (\(\text{D}_\text{3dgs}^\text{random}\)) and the proposed edge-based initialization (\(\text{D}_\text{3dgs}^\text{edge}\)). The results, presented in Table~\ref{tab:rebuttal-ablation}, indicate a significant performance drop when \( \mathcal{L}_{\text{nerf}} \) is removed, demonstrating that jointly optimizing the NeRF branch benefits the GS branch. Similarly, direct optimization of GS after initialization leads to performance degradation, validating the effectiveness of our proposed joint optimization strategy. Moreover, we observe that \(\text{D}_\text{3dgs}^\text{random}\) underperforms compared to \(\text{D}_\text{3dgs}^\text{edge}\), further confirming the superiority of our initialization strategy.  

\section{Per-scene Breakdown Results of NeRF-GS}
We provide a detailed quantitative assessment of NeRF-GS across various scenes in Tables~\ref{tab:per-blender},~\ref{tab:per-tank-deep} and~\ref{tab:per-mip360}, including metrics such as PSNR, SSIM, and LPIPS.

\begin{table}[h]
\centering
\caption{\textbf{Per-scene results of Blender dataset of our method}.}
\label{tab:per-blender}
\resizebox{1.\linewidth}{!}{
\begin{tabular}{c|ccccccccc}
\toprule
\multirow{2}[4]{*}{} & \multicolumn{9}{c}{Full views} \\
\cmidrule{2-10}      & chair & drums & ficus & hotdog & lego  & materials & mic   & ship  & Avg \\
\midrule
PSNR  & 35.36 & 26.34 & 35.15 & 37.81 & 36.45 & 30.873 & 36.78 & 30.9  & 33.71 \\
SSIM  & 0.985 & 0.948 & 0.9852 & 0.984 & 0.983 & 0.962 & 0.988 & 0.887 & 0.970 \\
LPIPS & 0.012 & 0.047 & 0.013 & 0.019 & 0.014 & 0.036 & 0.0075 & 0.111 & 0.032 \\
\midrule
\multirow{2}[4]{*}{} & \multicolumn{9}{c}{12 views} \\
\cmidrule{2-10}      & chair & drums & ficus & hotdog & lego  & materials & mic   & ship  & Avg \\
\midrule
PSNR  & 28.32 & 22.67 & 26.48 & 29.58 & 26.18 & 24.26 & 29.02 & 24.21 & 26.34 \\
SSIM  & 0.950  & 0.8991 & 0.9371 & 0.942 & 0.912 & 0.888 & 0.966 & 0.799 & 0.912 \\
LPIPS & 0.040  & 0.082 & 0.035 & 0.063 & 0.081 & 0.106 & 0.027 & 0.203 & 0.080 \\
\midrule
\multirow{2}[4]{*}{} & \multicolumn{9}{c}{8 views} \\
\cmidrule{2-10}      & chair & drums & ficus & hotdog & lego  & materials & mic   & ship  & Avg \\
\midrule
PSNR  & 25.95  & 20.58 & 23.12 & 27.27 & 25.01 & 20.83 & 25.72 & 22.93 & 23.92 \\
SSIM  & 0.917  & 0.871 & 0.892 & 0.937 & 0.885 & 0.834 & 0.941 & 0.773 & 0.881 \\
LPIPS & 0.061 & 0.114 & 0.101 & 0.099 & 0.101 & 0.184 & 0.112 & 0.225 & 0.124 \\
\bottomrule
\end{tabular}%

}
\end{table}

\begin{table}[h]
\centering
\caption{\textbf{Per-scene results of Tanks\&Temples and DeepBlending datasets of our method}.}
\label{tab:per-tank-deep}
\resizebox{0.8\linewidth}{!}{
\begin{tabular}{c|ccc|ccc}
\toprule
\multirow{2}[4]{*}{} & \multicolumn{3}{c|}{Tanks\&Temples} & \multicolumn{3}{c}{DeepBlending} \\
\cmidrule{2-7}      & Truck & Train & Avg   & Dr Johnson & Playroom & Avg \\
\midrule
PSNR  & 26.27 & 22.61 & 24.44 & 30.17 & 31.23 & 30.70 \\
SSIM  & 0.887 & 0.833 & 0.860 & 0.91  & 0.914 & 0.912 \\
LPIPS & 0.127 & 0.195 & 0.161 & 0.235 & 0.238 & 0.237 \\
\bottomrule
\end{tabular}%

}
\end{table}

\begin{table}[h]
\centering
\caption{\textbf{Per-scene results of Mip-NeRF360 dataset of our method}.}
\label{tab:per-mip360}
\resizebox{1.\linewidth}{!}{
\begin{tabular}{c|cccccccccc}
\toprule
      & bicycle & bonsai & counter & graden & kitchen & room  & stump & flowers & treehill & Avg \\
\midrule
PSNR  & 25.52 & 33.97 & 30.5  & 27.84 & 32.56 & 32.78 & 27.08 & 21.71 & 22.99 & 28.32 \\
SSIM  & 0.695 & 0.957 & 0.93  & 0.868 & 0.939 & 0.941 & 0.785 & 0.613 & 0.626 & 0.817 \\
LPIPS & 0.327 & 0.145 & 0.144 & 0.102 & 0.102 & 0.155 & 0.206 & 0.314 & 0.395 & 0.210 \\
\bottomrule
\end{tabular}%

}
\end{table}

\end{document}